\documentclass{llncs}

\let\llncssubparagraph\subparagraph
\let\subparagraph\paragraph
\usepackage[compact]{titlesec}
\let\subparagraph\llncssubparagraph
\usepackage{graphicx}
\usepackage{algorithm}
\usepackage{algorithmic}
\usepackage{setspace}
\usepackage{amsmath}
\usepackage{booktabs}
\usepackage{subfig}
\usepackage{rotating}
\usepackage{multirow}
\usepackage{threeparttable}

\newcommand{\etal}{\textit{et al.}}
\begin{document}
\title{Data-Free Adversarial Perturbations for Practical Black-Box Attack}
%
%
\author{Zhaoxin Huan\inst{1,2} \and Yulong Wang\inst{2,3} \and Xiaolu Zhang\inst{2} \and Lin Shang\inst{1} \and Chilin Fu\inst{2} \and Jun Zhou\inst{2}}
\authorrunning{Z. Author et al.}
%
\institute{Department of Computer Science and Technology, \\
State Key Laboratory for Novel Software Technology Nanjing University\\
\email{huanzhx@smail.nju.edu.cn}, \email{shanglin@nju.edu.cn}\\
\and
Ant Financial Services Group\\
\email{\{yueyin.zxl,chilin.fcl,jun.zhoujun\}@antfin.com}
\and
Department of Computer Science and Technology, Tsinghua university\\
\email{wang-yl15@mails.tsinghua.edu.cn}}
\maketitle              
\begin{abstract}
Neural networks are vulnerable to adversarial examples, which are malicious inputs crafted to fool pre-trained models. Adversarial
examples often exhibit black-box attacking transferability, which allows that adversarial examples crafted for one model can fool another model.
However, existing black-box attack methods require samples from the training data distribution to improve the transferability of adversarial examples across different models. Because of the
data dependence, the fooling ability of adversarial perturbations is only applicable when training data are accessible. In this paper, we present a data-free method for crafting adversarial perturbations that can fool a target model without any knowledge about the training data distribution. In the practical setting of a black-box attack scenario where attackers do not have access to target models and training data, our method achieves high fooling rates on target models and outperforms other universal adversarial perturbation methods. Our method empirically shows that current deep learning models are still at risk even when the attackers do not have access to training data.

\end{abstract}
\section{Introduction}
In recent years, deep learning models demonstrate impressive performance on various machine learning tasks~\cite{DBLP:conf/cvpr/DengDSLL009,He_2016_CVPR,huang2017densely}. However, recent works show that deep neural networks are
highly vulnerable to adversarial perturbations~\cite{DBLP:journals/corr/GoodfellowSS14,DBLP:journals/corr/SzegedyZSBEGF13}. Adversarial examples are small, imperceptible perturbations crafted to fool target models. The inherent weakness of lacking robustness to adversarial examples for deep neural networks brings out security concerns,
especially for security-sensitive applications which require strong reliability~\cite{DBLP:conf/iclr/KurakinGB17a}.
\\ \indent
With the knowledge of the structure and parameters
of a given model, many methods can successfully generate
adversarial examples in the white-box manner~\cite{DBLP:journals/corr/SzegedyZSBEGF13,DBLP:journals/corr/GoodfellowSS14}. A more severe issue is that adversarial examples can be transferred across different models, known as black-box attack~\cite{DBLP:journals/corr/GoodfellowSS14}. This transferability allows for adversarial attacks without the knowledge of the structure and parameters of the target model. Existing black-box attack methods focus on improving the transferability of adversarial examples across different models under the assumption that attackers can obtain the training data on which the target models are trained~\cite{DBLP:conf/cvpr/DongLPS0HL18,DBLP:journals/corr/GoodfellowSS14,Huang_2019_ICCV}. Attackers firstly train a substitute model on the same training data, and then generate adversarial examples in the white-box manner. The perturbations crafted for substitute model can thus fool target model, since different models learn similar decision boundaries on the same training set~\cite{DBLP:journals/corr/GoodfellowSS14,Huang_2019_ICCV}.
\\ \indent
In practice, however, attackers can hardly obtain the training data for target model, even the number of categories. For example, the Google Cloud Vision API2 (GCV) only outputs scores for a number of top classes. On this real-world black-box setting, most of existing black-box attack methods can not be applied.
\\ \indent
In this paper, we present a data-free approach for crafting adversarial perturbations to address the above issues. Our method is to craft data-free perturbations that can fool the target model without any knowledge about the data distribution (e.g., the number of categories, type of data, etc.). We utilize such a property that the features extracted from different models are usually similar, since most models are fine-tuned from common pre-trained model weights~\cite{DBLP:journals/corr/abs-1801-05746}.
Therefore, we establish a mapping connection between fine-tuned model and pre-trained model. Instead of optimizing an objective that reduces the score of the predicted labels~\cite{DBLP:journals/corr/GoodfellowSS14,DBLP:conf/cvpr/DongLPS0HL18}, we propose to learn adversarial perturbations that can disturb the internal representation. Our proposed attack method views the logit outputs of pre-trained model as the extracted internal representation, and iteratively maximizes the divergence between clean images and their adversarial examples measured in this representation space. Because of the mapping connection, pre-trained model and fine-tuned model are similar in the internal representation and adversarial examples will successfully mislead target model with high probability.
\\ \indent
We evaluate the proposed method on two public datasets: CIFAR-10~\cite{krizhevsky2009learning} and Caltech-101~\cite{DBLP:journals/cviu/Fei-FeiFP07} and one private dataset with various models including state-of-the-art classifiers (e.g., ResNet~\cite{DBLP:journals/corr/SimonyanZ14a}, DenseNet~\cite{huang2017densely}, etc.). Experimental results show that on the real-world black-box setting, our method achieves significant attacking success rates.
In this practical setting of black-box attack scenario, only universal adversarial perturbation methods can be applied since they are image-agnostic methods.
Compared with universal adversarial perturbations (UAP)~\cite{DBLP:conf/cvpr/Moosavi-Dezfooli17} and generalizable data-free universal adversarial perturbations (GD-UAP)~\cite{DBLP:journals/pami/MopuriGB19}, the proposed method has the following advantages. First, our method outperforms UAP and GD-UAP by 8.05\% and 6.00\%. Second, UAP requires a number of training samples to converge when crafting an image-agnostic perturbation and GD-UAP also need to know the distribution of training data to achieve better performance. In contrast, our method generates adversarial perturbations without knowing the data distribution. Third, the proposed method does not need training phase. The perturbation can be obtained by a single back-propagation, whereas UAP and GD-UAP need to train universal perturbation until it converges.


\section{Related Work}\label{sec:related-work}
\subsubsection{White-Box Attack}
With the knowledge of the structure and parameters
of a given model, many methods can successfully generate adversarial examples in the white-box manner. Most white-box algorithms generate adversarial examples based on the gradient of loss function with
respect to the inputs. Szegedy\etal~\cite{DBLP:journals/corr/SzegedyZSBEGF13} first introduce adversarial examples generation by analyzing the instability of deep neural networks. Goodfellow\etal~\cite{DBLP:journals/corr/GoodfellowSS14}  further explain the phenomenon of adversarial examples
by analyzing the linear behavior of deep neural network and propose a simple and efficient adversarial examples generating method. Recently, Yinpeng Dong\etal~\cite{DBLP:conf/cvpr/DongLPS0HL18} integrate the momentum term into the iterative process for fast gradient sign to achieve better attack performance.

\subsubsection{Black-Box Attack}
The existing black-box attacks can be classified as query-based and transfer-based. In query-based methods, the attacker iteratively queries the outputs of target model and estimates the gradient of target model~\cite{DBLP:conf/ccs/ChenZSYH17}.
 As for transfer-based methods, the existing methods mainly focus on improving the transferability of adversarial examples across different models~\cite{Huang_2019_ICCV}. They assume the adversary can obtain the training data without
the knowledge of the structure and parameters of target model. Because query-based method requires thousands of queries, it is hard to be used in practical attack. In this paper, we focus on transfer-based black-box attack.

Recent work by Moosavi-Dezfooli \etal~\cite{DBLP:conf/cvpr/Moosavi-Dezfooli17} presents the existence of image-agnostic perturbations,
called universal adversarial perturbations (UAP) that can fool the state-of-the-art recognition models on most clean images.
Mopuri\etal~\cite{DBLP:journals/pami/MopuriGB19} further proposed a generalizable approach for crafting universal adversarial perturbations, called generalizable data-free universal adversarial perturbations (GD-UAP). These two image-agnostic universal adversarial perturbations can effectively attack under real-world black-box setting. Instead of seeking universal adversarial perturbations, our method is to generate image-specific perturbations without knowing data distribution.

\section{Data-Free Adversarial Perturbations}\label{sec:method}
Based on the motivation presented in the introduction, we propose the data-free attack framework. We combine the idea of feature-level attack with the mapping connection between fine-tuned model and pre-trained model to facilitate black-box attacking on target model without knowing the data distribution. Specifically, we use the output of pre-trained model as internal representation to measure the difference between clean image and adversarial example. By iteratively maximizing the divergence with respect to our objective function Equation~\eqref{eq1}, the internal representation becomes much more different. Finally, because of the mapping connection, adversarial examples will successfully mislead target model with high probability. We briefly show our attack framework in Algorithm~\ref{alg:attack algorithm}.

\begin{algorithm}[htb]
\caption{Data-free adversarial attack algorithm.}
\label{alg:attack algorithm}
\begin{algorithmic}[1]
\REQUIRE~~\\
A clean image $x$; \\
The target model $f(x)$; \\
The pre-trained model $t(x)$;
\ENSURE~~\\
The adversarial perturbations $x^{*}$ which misleads target model $f(x)$.
\STATE
Initialize $x^{*}$ with $x$;
\STATE
Compute the objective function Equation~\eqref{eq1} with respect to $t(x)$ for $x$;
\STATE
Use numerical optimization to iteratively maximize the divergence between $x$ and $x^{*}$ by Equation~\eqref{eq1};
\STATE
Get the adversarial perturbations $x^{*}$ generated by $t(x)$;
\STATE
$x^{*}$ misleads target model $f(x)$;

\end{algorithmic}
\end{algorithm}

\subsection{Problem Definition}\label{sec:problem definition}
Let $x$ denote the clean image from a given dataset, and $y_{true}$ denote the class. A target  model is a function $f(x)=y$ that accepts an input $x\in X$ and and produces an output $y\in Y$.
$f(x)$ is the outputs of target model including the softmax function, define $f_{l}(x)=z$ to be the output of final layer before the softmax output($z$ are also called logits), and $f(x)=softmax(f_{l}(x))=y$. The goal of adversarial attack is to seek an example $x^{*}$ with the magnitude of adversarial perturbation $\epsilon$ which is misclassified by the target model.

\subsection{Black-Box Setting}\label{sec:black-box setting}
In this paper, we use the definition of real-world black-box: the adversary can not obtain the structure and parameters of target model as well as its data distribution (e.g., the number
of categories, type of data, etc.). Moreover, the target model is fine-tuned on pre-trained model.
Let $t(x):x\in X^{'} \rightarrow y\in Y{'}$ denote the pre-trained model, where $X^{'} \neq X, Y^{'} \neq Y$. Our objective is to establish a mapping connection between $f(x)$ and $t(x)$ and utilize $t(x)$ to craft data-free perturbations that can fool $f(x)$.

\subsection{Mapping Connection}\label{sec:mapping connection}
\begin{figure}[thp]
\includegraphics[width=\textwidth]{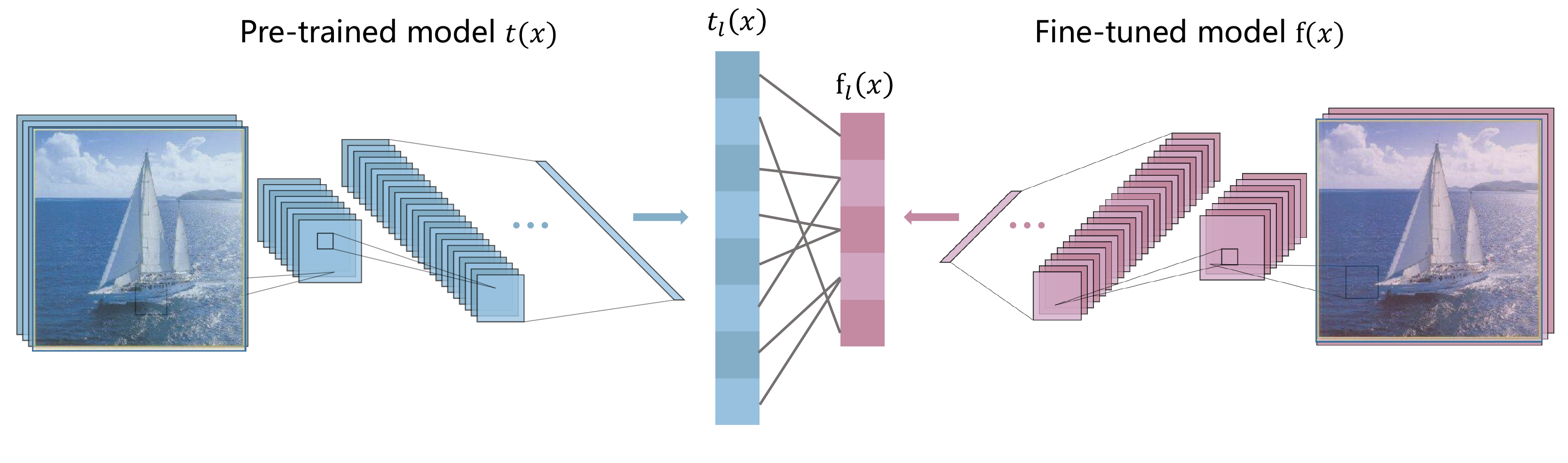}
\caption{Mapping connection between fine-tuned model $f(x)$ and pre-trained model $t(x)$. Logits output $f_{l}(x)$ and $t_{l}(x)$ may contain mapping relationship.} \label{fig1}
\end{figure}

For general image classification tasks, the extracted features are similar. Instead of initializing the model with random weights, initializing it with the pre-trained model can boost performance and reduce training time.
Therefore, it is common to use pre-trained model to fine-tune on new tasks~\cite{DBLP:journals/corr/abs-1801-05746}. In this paper, we establish the relationship between fine-tuned model $f(x)$ and pre-trained model $t(x)$, called mapping connection. As shown in Figure~\ref{fig1}, even though the training data distribution between $f(x)$ and $t(x)$ is different ($X \neq X^{'}$), we consider the logits output between these two models contain the 'mapping' connection: given an input $x$, each neuron in $f_{l}(x)$ may be obtained by weighted summation of neurons in $t_{l}(x)$. We will give some experimental explanations in Section~\ref{sec:Experimental phenomenon of mapping connection}.
Therefore, by generating the adversarial perturbations from $t(x)$, it will successfully mislead $f(x)$ with high probability.

\subsection{Maximizing Divergence}
Note in Section~\ref{sec:problem definition} and Section~\ref{sec:black-box setting}, $f(x):x\in X \rightarrow y\in Y$ is target model and $t(x):x\in X^{'} \rightarrow y\in Y{'}$ is pre-trained model. Because $f(x)$ is fine-tuned from $t(x)$, the data distribution is different from each other ($X  \neq X^{'}$ and $Y  \neq Y^{'}$). Given a clean image $x \in X$, our goal is to utilize $t(x)$ to generate corresponding adversarial example $x^{*}$ which can mislead target model as $f(x^{*})\neq y_{true}$.

Our objective is to craft data-free perturbations that can fool the target model without any knowledge about the data distribution (e.g., the number of categories, type of data, etc.). Therefore, instead of optimizing an objective that reduces the score to the predicted label or flip the predicted label~\cite{DBLP:journals/corr/GoodfellowSS14,DBLP:conf/cvpr/DongLPS0HL18}, we propose to learn perturbations that can maximize feature divergence between clean images and adversarial examples.

More precisely, $x^{*}$ is formulated as a constrained optimization problem:
\begin{gather}
x^{*}=\arg\max_{x^{'}}\left \| \left | t_{l}(x) \right | * \frac{t_{l}(x)}{t_{l}(x^{'})} \right \|_{2}^{2} \notag \\
subject \ to \left \| x - x ^{*} \right \|_{\infty } < \epsilon
\label{eq1}
\end{gather}
\noindent
where $t_{l}(x)$ is the output at logits (pre-softmax) layer. Equation~\eqref{eq1} measures the divergence between the logits output of $x$ and $x^{'}$. $|t_{l}(x)|$ represents magnitude of each element in $t_{l}(x)$ and $\frac{t_{l}(x)}{t_{l}(x^{'})}$ represents the difference between $t_{l}(x)$ and $t_{l}(x^{'})$. Intuitively, our objective function in Equation~\eqref{eq1} increases or decreases $t_{l}(x^{'})$ according to the direction of $t_{l}(x)$. And the magnitude of the change depends on weight $|t_{l}(x)|$. We will show the effectiveness of our objective function in Section~\ref{sec:The effectiveness of objective function}. The constraint on the distance between $x$ and $x^{*}$ is formulated in terms of the
$L_{\infty}$ norm to limit the maximum deviation of single pixel to $\epsilon$. The goal is to constrain the degree to which the
perturbation is perceptible.

Previous adversarial examples generation methods~\cite{DBLP:journals/corr/SzegedyZSBEGF13,DBLP:journals/corr/GoodfellowSS14,DBLP:conf/cvpr/DongLPS0HL18} aim to increase the loss function according to the gradient of $f(x)$ (softmax output). However, due to the deep hierarchy of architectures, the gradients of loss with respect to input may vanish during propagation. To address this issue,
we aim to maximize the divergence of logits output $f_{l}(x)$
between input $x$ and adversarial example $x^{*}$. Empirically, we found that it is inappropriate to directly use objective functions such as Kullback–Leibler divergence to measure the divergence, since the optimization can be hard to converge.

\subsection{Implementation Details}
For implementation details, we first scale the input x into $[-1,1]$ and initialize $x^{'}=x$. Then, we compute the gradients of objective~\eqref{eq1} with respect to input $x$. The adversarial examples will be updated by multiple steps. In each step, we take the sign function of the gradients and clip the adversarial examples into $[-1,1]$ to make valid images. Algorithm~\ref{alg:Numerical optimization} presents the details of perturbations generation.

\begin{algorithm}[htb]
\caption{Implementation details for data-free perturbations.}
\label{alg:Numerical optimization}
\begin{algorithmic}[1]
\REQUIRE~~\\
The clean image $x$; \\
The  maximum deviation of single pixel $\epsilon$;\\
The number of iterations $n$;
\ENSURE~~\\
The adversarial perturbations $x^{*}$ generated by pre-trained model $t(x)$;
\STATE Initialize: $x^{'}=x, \epsilon^{'} = \frac{\epsilon}{n}, i=0$;

\WHILE {$i < n$}
\STATE Maximize divergence between $x$ and $x^{'}$ by Equation~\eqref{eq1}:\\
$x^{'}_{i+1} = clip(x^{'}_{i}+\epsilon^{'}sign(\bigtriangledown_{x}\left \| \left | t_{l}(x) \right | * \frac{t_{l}(x)}{t_{l}(x^{'})} \right \|_{2}^{2} ),-1,1)$
\ENDWHILE
\RETURN $x^{*} = clip(x+\epsilon sign(x-x^{'}_{n}),-1,1)$;

\end{algorithmic}
\end{algorithm}

\section{Experiments}\label{sec:experiments}
In this section, we present the experimental results to demonstrate the effectiveness of our data-free adversarial perturbation method.

\subsection{Experimental Settings}\label{sec:Experimental settings}
Throughout experiments, target models are fine-tuned based on ImageNet~\cite{DBLP:conf/cvpr/DengDSLL009} pre-trained models. We first fine-tune the target model on different datasets to simulate a practical training scenario. Then we only use pre-trained models to generate adversarial perturbations without knowing the training data distribution or the architecture of target models by Algorithm~\ref{alg:attack algorithm}.

We explore four mainstream deep models: GoogleNet~\cite{DBLP:conf/cvpr/SzegedyLJSRAEVR15}, VGG-16~\cite{DBLP:journals/corr/SimonyanZ14a}, ResNet-152~\cite{He_2016_CVPR} and DenseNet-169~\cite{huang2017densely}. We compare our method to UAP ~\cite{DBLP:conf/cvpr/Moosavi-Dezfooli17} and GD-UAP~\cite{DBLP:journals/pami/MopuriGB19}. Although some classic attack algorithms such as FGSM~\cite{DBLP:journals/corr/GoodfellowSS14} and MI-FGSM~\cite{DBLP:conf/cvpr/DongLPS0HL18} are data dependence which are not directly comparable to ours, we evaluate their attack performance under this black-box attack scenario. For all the following experiments, perturbation crafted by our method is termed as $DFP$. The maximum perturbation $\epsilon$ is set to 10 among all experiments, with pixel value in $[0, 255]$, and the number of iterations is 10.

\subsection{Datasets}\label{sec:dataset}
\subsubsection{CIFAR-10}
The CIFAR-10 dataset~\cite{krizhevsky2009learning} consists of 60,000 colour images across 10 classes, with size of $32\times32$. We use training images to fine-tune target models which are pre-trained on ImageNet and use test images to evaluate attack performance. Since UAP and GD-UAP are high resolution perturbations (usually in $224\times224$), directly using low-resolution images from CIFAR-10 is inappropriate. Before fine-tuning target models, we resize images to $224\times224$ without losing recognition performance.
\subsubsection{Caltech101}
Caltech101~\cite{DBLP:journals/cviu/Fei-FeiFP07}consists of objects belonging to 101 categories. The size of each image is roughly $300\times200$ pixels. Compared with CIFAR-10, Caltech101 is more complicated with higher resolution.
\subsubsection{Cosmetic Insurance Dataset}
To fully illustrate the effectiveness of our attack method, we construct another private real-world dataset, called cosmetic insurance dataset. This dataset consists of credentials for customer who are allergic to cosmetic products, including cosmetic products, allergic skin, medical record, etc. This dataset does not involve any Personal Identifiable Information (PPI). The data is only used for academic research and processed by sampling. During the experiment, we conduct adequate data protection to prevent the risk of data leakage and destroy it after the experiment.

\subsection{Data-Free Attack Ability} \label{sec:Data-Free attack ability}
Table~\ref{tab:1} presents the attack performance achieved by our objective on various network architectures on three datasets. \textit{Baseline} means the model's error rate on clean image (without perturbation).
\textit{Fooling rate} is the percentage of test images for which our crafted perturbations successfully alter the predicted label. Each row in the table indicates one target model and the columns indicate different attack methods. Since UAP and GD-UAP do not provide the perturbation on Densenet-169, we use  ``$\backslash$'' in the table. Our perturbations result in an average fooling rate of 29.23\% on Caltech101 which is 8.05\% and 6.00\% higher than UAP and GD-UAP. Moreover, compared with UAP and GD-UAP, our method crafts perturbation by one single back propagation without knowing any training data distribution which is much more efficient in practical scenario.

\begin{table}[!t]
  \centering
  \caption{Data-free attack results on CIFAR-10, Caltech101 and cosmetic insurance datasets. Each row in the table shows fooling rates (\%) for perturbation generated by different attack methods when attacking various target models (columns). Fooling rate is the percentage of test images for which the crafted perturbations successfully alter the predicted label. Baseline in table means the model’s error rate on clean images}
  \scalebox{0.9}{ 
  \renewcommand\tabcolsep{2.9pt} 
  \begin{threeparttable}
    \begin{tabular}{clcccccc}
    \toprule
    \multicolumn{2}{c}{\textbf{Model}} & Baseline
    & GD-UAP~\cite{DBLP:journals/pami/MopuriGB19}
    & UAP~\cite{DBLP:conf/cvpr/Moosavi-Dezfooli17}
    & FGSM~\cite{DBLP:journals/corr/GoodfellowSS14}
    & MIFGSM~\cite{DBLP:conf/cvpr/DongLPS0HL18}
    & \textbf{DFP}
    \\
    \midrule
    \multirow{3}[1]{*}{{\begin{sideways}CIFAR-10\end{sideways}}}
          & GoogleNet & 10.08  & 18.81  & 14.13  & 11.01 & 12.32&\textbf{25.26}\\
          & VGG-16 & 9.81  & 17.23  & 13.21  & 10.23 &11.57 &\textbf{24.37}\\
          & ResNet-152 & 8.23  & 18.09  & 15.12  & 9.89 &10.28 &\textbf{28.73}\\
          & DenseNet-169 & 8.05  & $\backslash$  & $\backslash$  & 8.97 &10.00 &\textbf{25.64}\\
    \midrule
    \multirow{3}[1]{*}{{\begin{sideways}Caltech101\end{sideways}}}
    & GoogleNet & 15.31  & 23.41  & 21.16  & 16.21 & 16.68&\textbf{28.47}\\
          & VGG-16 & 15.27  & 24.17  & 22.00  & 15.76 &16.00 &\textbf{29.61}\\
          & ResNet-152 & 13.71  & 22.13  & 20.38  & 14.58 &15.86 &\textbf{28.77}\\
          & DenseNet-169 & 13.68  & $\backslash$  & $\backslash$  & 15.01 &10.00 &\textbf{30.09}\\
    \midrule
    \multirow{4}[1]{*}{{\begin{sideways}cosmetic\end{sideways}}}
    & GoogleNet & 13.28  & 18.78  & 16.53  & 14.27 & 15.25&\textbf{22.73}\\
          & VGG-16 & 12.69  & 17.60  & 15.27  & 13.83 &14.77 &\textbf{22.41}\\
          & ResNet-152 & 10.63  & 18.97  & 14.49  & 12.21 &13.39 &\textbf{23.01}\\
          & DenseNet-169 & 8.43  & $\backslash$  & $\backslash$  & 10.93 &11.01 &\textbf{21.84}\\
    \midrule
    \end{tabular}%
    \end{threeparttable}
  }
  \label{tab:1}%
\end{table}%

Although previous attack methods such as FGSM~\cite{DBLP:journals/corr/GoodfellowSS14} and MI-FGSM~\cite{DBLP:conf/cvpr/DongLPS0HL18} are training-data dependent which are not directly comparable to ours, we evaluate their attack performance under this black-box attack scenario shown in Table~\ref{tab:1}. It is clear that the fooling rates of FGSM and MI-FGSM under this practical scenario are significant lower than $DFP$. Because of the data dependence, fooling ability of the crafted perturbations in FGSM and MI-FGSM is limited to the available training data.

\begin{figure}[htp]
\includegraphics[width=\textwidth]{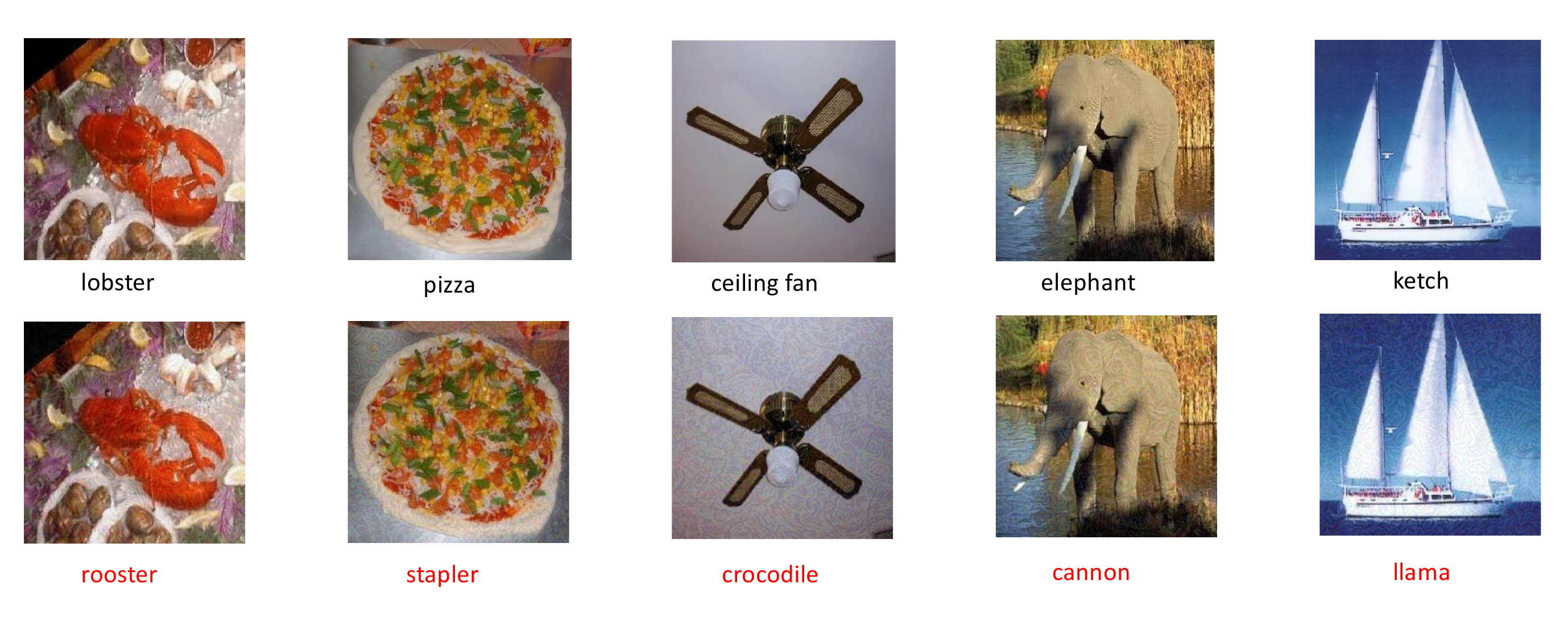}
\caption{Original and adversarial image pairs from Caltech101 dataset generated for ResNet. First row shows original images and
corresponding predicted labels, second row shows the corresponding
perturbed images with their wrong predictions.} \label{fig2}
\end{figure}

Figure~\ref{fig2} shows example data-free perturbations crafted by the proposed method. The top row shows the clean and bottom row shows the corresponding adversarial images. The perturbed images are visually indistinguishable form their corresponding clean images. All the clean images shown in the figure are correctly classified and are successfully fooled by the added perturbation. Corresponding label predicted by the model is shown below each image. The correct labels are shown in black color and the wrong ones in red.

\subsection{Black-Box Attack Transferability}\label{sec:Transfer ability across models}
In section~\ref{sec:Data-Free attack ability}, we report the attack performance without knowing the training data distribution. In this section, we evaluate the fooling rates of black-box attack across different models. Each row in the Table~\ref{tab:2} indicates the target model which generates perturbations and the columns indicate various models attacked using the learned perturbations. The diagonal fooling rates indicate the \textit{data-free white-box} attack noted in Section~\ref{sec:Data-Free attack ability}, where all the information about the model is known to the attacker except training data distribution. The off-diagonal rates indicate \textit{real-world black-box} attack, where no information about the model's architecture or training data distribution under attack is revealed to the attacker. Our perturbations cause a mean
white-box fooling rate of 25.91\% and a mean black-box fooling rate of 15.04\%. Given the data-free nature of the optimization, these fooling rates are alarmingly significant.

\begin{table}[!t]
  \centering
  \caption{The transferability of our attack method on CIFAR-10, Caltech101 and cosmetic insurance datasets. Each row in the table shows fooling rates (\%) for perturbation learned on a specific target model when attacking various other models (columns). Diagonal rates indicate data-free white-box attack and off-diagonal rates represent real-world black-box attack scenario.}
  \scalebox{1.0}{ 
  \renewcommand\tabcolsep{2.9pt} 
  \begin{threeparttable}
    \begin{tabular}{clcccc}
    \toprule
    \multicolumn{2}{c}{\textbf{Model}} & GoogleNet
    & VGG-16
    & ResNet-152
    & DenseNet-169
    \\
    \midrule
    \multirow{3}[1]{*}{{\begin{sideways}CIFAR-10\end{sideways}}}
          & GoogleNet & 25.26&  15.37& 12.23 &12.47 \\
          & VGG-16 & 13.83& 24.37 &11.63  &11.50 \\
          & ResNet-152 &13.43 &14.41  & 28.73&19.65\\
          & Densenet-169 & 14.01 &14.65 &17.72 & 25.64\\
    \midrule
    \multirow{3}[1]{*}{{\begin{sideways}Caltech101\end{sideways}}}
    & GoogleNet & 28.47&  17.23&  14.40&15.21 \\
          & VGG-16 &18.88 & 29.61 & 16.63 &16.01 \\
          & ResNet-152 &15.53 & 16.98 & 28.77& 17.75\\
          & DenseNet-169 & 14.29& 14.71& 17.26 & 30.09\\
    \midrule
    \multirow{4}[1]{*}{{\begin{sideways}cosmetic\end{sideways}}}
    & GoogleNet &  22.73&  15.20&  14.17&14.48 \\
          & VGG-16 &14.49 &22.41  &13.67  & 13.88 \\
          & ResNet-152 & 14.40& 13.97 &23.01 & 17.08\\
          & DenseNet-169  & 15.21& 13.59 & 15.63 &21.84\\
    \midrule
    \end{tabular}%
    \end{threeparttable}
  }
  \label{tab:2}%
\end{table}%

\subsection{Empirical Demonstration of Mapping Connection
}\label{sec:Experimental phenomenon of mapping connection}
\begin{figure}[t]
\centering
\includegraphics[width=0.6\textwidth,height=0.4\textwidth]{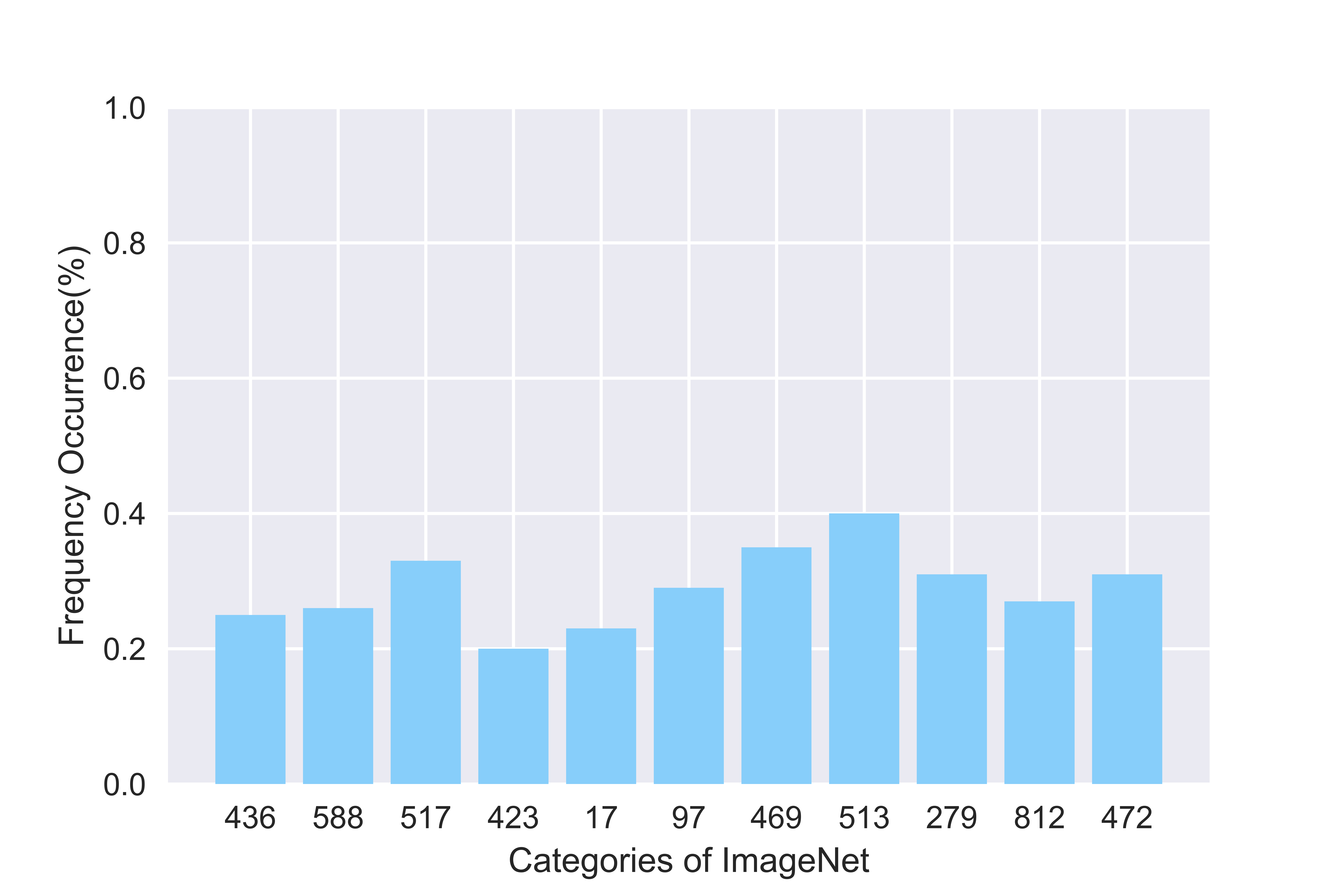}
\caption{Experimental explanation of mapping connection. The horizontal axis represents categories in ImageNet, and vertical axis represents the proportion of test images in the cosmetic insurance dataset that is classified as categories in horizontal axis.} \label{fig4}
\end{figure}

As a further analysis, we reveal the mapping connection between fine-tuned model and pre-trained model noted in Section~\ref{sec:mapping connection}.
Since the categories of cosmetic insurance dataset have no overlap with ImangeNet~\cite{DBLP:journals/cviu/Fei-FeiFP07}, we evaluate test images from cosmetic insurance dataset with ImageNet pre-trained DenseNet-169 and calculate the frequency occurrence shown in Figure~\ref{fig4}. The horizontal axis represents categories in ImageNet, and vertical axis represents the proportion of test images in the cosmetic insurance dataset that is classified as categories in horizontal axis. For example, by evaluating test images belonging to chat record category, there are 35\% images classified as category ``caldron''  in ImageNet, which has no relationship with chat record.

The frequency occurrence of each category in Figure~\ref{fig4} is higher than 20\%. This phenomenon demonstrates that even though fine-tuned model has different categories of pre-trained model, the logits outputs between these two models still contain relationship. Therefore, by disturbing the logits outputs of pre-trained model, it will successfully disturb the logits output of target model with high probability, which can cause the wrong prediction.

\subsection{Effectiveness of Objective Function}\label{sec:The effectiveness of objective function}

\begin{figure}[thb]
    \begin{minipage}[t]{0.5\linewidth}
    \centering
    \includegraphics[height=0.6\textwidth,width=0.8\textwidth]{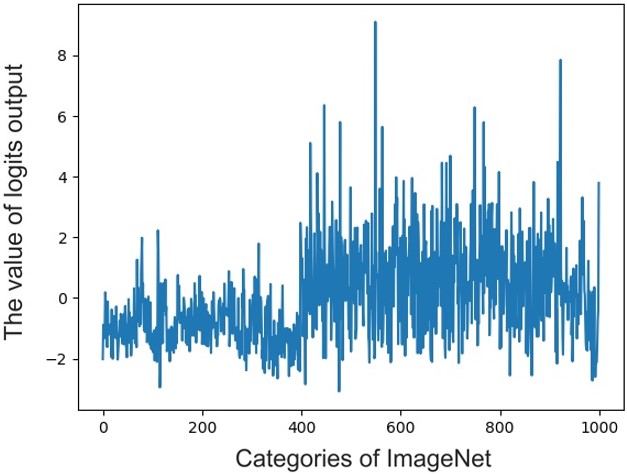}
    \end{minipage}
    \begin{minipage}[t]{0.5\linewidth}
    \centering
    \includegraphics[height=0.6\textwidth,width=0.8\textwidth]{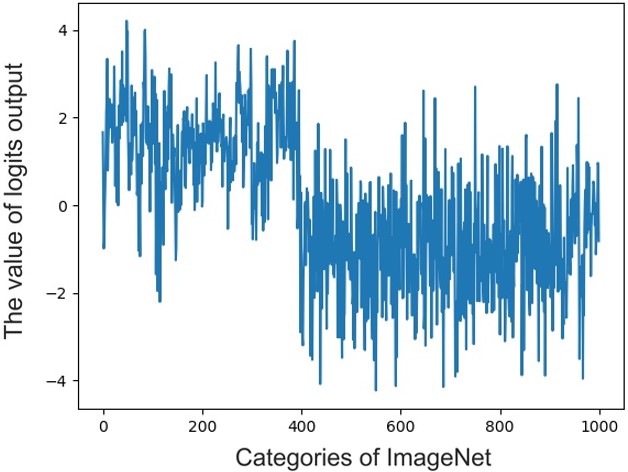}
    \end{minipage}
    \caption{The left image is $t_{l}(x)$ and right image is $t_{l}(x^{*})$. The horizontal axis represents each category in ImageNet ($t_{l}(x)_{i}, i = 1,2,\cdots,1000$), and vertical axis represents the value of logits.}\label{fig3}
\end{figure}

To demonstrate the effectiveness of our objective function~\eqref{eq1}, we compare the logits outputs between a clean image $t_{l}(x)$ (left) and the corresponding adversarial example $t_{l}(x^{*})$ (right) after optimizing Equation~\eqref{eq1}, shown in Figure~\ref{fig3}. The horizontal axis represents each category in ImageNet ($t_{l}(x)_{i}, i = 1,2,\cdots,1000$), and vertical axis represents the value of logits. It can be seen from the figure that $t_{l}(x)$ and $t_{l}(x^{*})$ have a significant divergence in magnitude and direction. Combined with mapping connection, it make sense that our objective function dose craft effective data-free perturbations illustrated in Section~\ref{sec:Data-Free attack ability} and Section~\ref{sec:Transfer ability across models}.



\section{Conclusion}\label{sec:conclusion}
In this paper, we have proposed a data-free objective to generate adversarial perturbations. Our objective is to craft data-free perturbations that can fool the target model without any knowledge about the data distribution (e.g., the number
of categories, type of data, etc.). Our method does not need to utilize any training data sample, and we propose to generate perturbations that can disturb the internal representation. Finally, we demonstrate that our objective of crafting data-free adversarial perturbations is effective to fool target model without knowing training data distribution or the architecture of models. The significant fooling rates achieved by our method emphasize that the current deep learning models are now at an increased risk.

{
\bibliographystyle{splncs04}
\bibliography{egbib}

\begin{thebibliography}{10}
\providecommand{\url}[1]{\texttt{#1}}
\providecommand{\urlprefix}{URL }
\providecommand{\doi}[1]{https://doi.org/#1}

\bibitem{DBLP:conf/ccs/ChenZSYH17}
Chen, P., Zhang, H., Sharma, Y., Yi, J., Hsieh, C.: {ZOO:} zeroth order
  optimization based black-box attacks to deep neural networks without training
  substitute models. In: Proceedings of the 10th {ACM} Workshop on Artificial
  Intelligence and Security (November 2017)

\bibitem{DBLP:conf/cvpr/DengDSLL009}
Deng, J., Dong, W., Socher, R., Li, L., Li, K., Li, F.: Imagenet: {A}
  large-scale hierarchical image database. In: The IEEE Conference on Computer
  Vision and Pattern Recognition (CVPR) (June 2009)

\bibitem{DBLP:conf/cvpr/DongLPS0HL18}
Dong, Y., Liao, F., Pang, T., Su, H., Zhu, J., Hu, X., Li, J.: Boosting
  adversarial attacks with momentum. In: The IEEE Conference on Computer Vision
  and Pattern Recognition (CVPR) (June 2018)

\bibitem{DBLP:journals/corr/GoodfellowSS14}
Goodfellow, I.J., Shlens, J., Szegedy, C.: Explaining and harnessing
  adversarial examples. In: The International Conference on Learning
  Representations (ICLR) (May 2015)

\bibitem{He_2016_CVPR}
He, K., Zhang, X., Ren, S., Sun, J.: Deep residual learning for image
  recognition. In: The IEEE Conference on Computer Vision and Pattern
  Recognition (CVPR) (June 2016)

\bibitem{huang2017densely}
Huang, G., Liu, Z., van~der Maaten, L., Weinberger, K.Q.: Densely connected
  convolutional networks. In: The IEEE Conference on Computer Vision and
  Pattern Recognition (CVPR) (July 2017)

\bibitem{Huang_2019_ICCV}
Huang, Q., Katsman, I., He, H., Gu, Z., Belongie, S., Lim, S.N.: Enhancing
  adversarial example transferability with an intermediate level attack. In:
  The IEEE International Conference on Computer Vision (ICCV) (October 2019)

\bibitem{DBLP:journals/corr/abs-1801-05746}
Iglovikov, V., Shvets, A.: Ternausnet: U-net with {VGG11} encoder pre-trained
  on imagenet for image segmentation. CoRR  \textbf{abs/1801.05746} (2018)

\bibitem{krizhevsky2009learning}
Krizhevsky, A., Hinton, G., et~al.: Learning multiple layers of features from
  tiny images. Tech. rep., Citeseer (2009)

\bibitem{DBLP:conf/iclr/KurakinGB17a}
Kurakin, A., Goodfellow, I.J., Bengio, S.: Adversarial examples in the physical
  world. In: The International Conference on Learning Representations (ICLR)
  (April 2017)

\bibitem{DBLP:journals/cviu/Fei-FeiFP07}
Li, F., Fergus, R., Perona, P.: Learning generative visual models from few
  training examples: An incremental bayesian approach tested on 101 object
  categories. Computer Vision and Image Understanding  (2007)

\bibitem{DBLP:conf/cvpr/Moosavi-Dezfooli17}
Moosavi{-}Dezfooli, S., Fawzi, A., Fawzi, O., Frossard, P.: Universal
  adversarial perturbations. In: The IEEE Conference on Computer Vision and
  Pattern Recognition (CVPR) (July 2017)

\bibitem{DBLP:journals/pami/MopuriGB19}
Mopuri, K.R., Ganeshan, A., Babu, R.V.: Generalizable data-free objective for
  crafting universal adversarial perturbations. {IEEE} Trans. Pattern Anal.
  Mach. Intell.  (2019)

\bibitem{DBLP:journals/corr/SimonyanZ14a}
Simonyan, K., Zisserman, A.: Very deep convolutional networks for large-scale
  image recognition. In: The International Conference on Learning
  Representations (ICLR) (May 2015)

\bibitem{DBLP:conf/cvpr/SzegedyLJSRAEVR15}
Szegedy, C., Liu, W., Jia, Y., Sermanet, P., Reed, S.E., Anguelov, D., Erhan,
  D., Vanhoucke, V., Rabinovich, A.: Going deeper with convolutions. In: The
  IEEE Conference on Computer Vision and Pattern Recognition (CVPR) (June 2015)

\bibitem{DBLP:journals/corr/SzegedyZSBEGF13}
Szegedy, C., Zaremba, W., Sutskever, I., Bruna, J., Erhan, D., Goodfellow,
  I.J., Fergus, R.: Intriguing properties of neural networks. In: The
  International Conference on Learning Representations (ICLR) (April 2014)

\end{thebibliography}
}
\end{document}